\pdfoutput=1

\documentclass[11pt]{article}

\usepackage{acl}

\usepackage{times}
\usepackage{latexsym}

\usepackage[T1]{fontenc}

\usepackage[utf8]{inputenc}

\usepackage{microtype}

\usepackage{todonotes}
\usepackage{multirow}

\title{From Stance to Concern: \\ Adaptation of Propositional Analysis to New Tasks and Domains}

\author{Brodie Mather \and  Bonnie Dorr \and Adam Dalton \and William de Beaumont \\ Institute for Human \& Machine Cognition\\
\AND
Owen Rambow \\ Stony Brook University \\
\AND
Sonja Schmer-Galunder\\ Smart Information Flow Technologies 
}

\date{}

\begin{document}
\maketitle
\begin{abstract}
We present a generalized paradigm for adaptation of propositional analysis (predicate-argument pairs) to new tasks and domains. We leverage an analogy between \textit{stances} (belief-driven sentiment) and \textit{concerns} (topical issues with moral dimensions/endorsements) to produce an \textit{explanatory} representation. A key contribution is the combination of semi-automatic resource building for extraction of domain-dependent concern types (with 2-4 hours of human labor per domain) and an entirely automatic procedure for extraction of domain-independent moral dimensions and endorsement values.  Prudent (automatic) selection of terms from propositional structures for lexical expansion (via semantic similarity) produces new moral dimension lexicons at three levels of granularity beyond a strong baseline lexicon. We develop a ground truth (GT) based on expert annotators and compare our concern detection output to GT, to yield 231\% improvement in recall over baseline, with only a 10\% loss in precision. F1 yields 66\% improvement over baseline and 97.8\% of human performance. Our lexically based approach yields large savings over approaches that employ costly human labor and model building. We provide to the community a newly expanded moral dimension/value lexicon, annotation guidelines, and GT.






\end{abstract}

\section{Introduction}
\label{sec:intro}

This paper presents a generalized paradigm for adaptation of tasks involving predicate-argument pairs, i.e., combinations of actions and their participants, to new tasks and domains. Predicate-argument analysis has been a longstanding area of research for many tasks: event detection \cite{du2020,zhang2020}, opinion extraction \cite{yang2013}, textual entailment \cite{stern2014}, and coreference \cite{shibata2018}. We refer to the induction of such representations as \textit{propositional analysis}. We induce a proposition PREDICATE(x$_1$,x$_2$,...) to represent sentences such as \textit{John wears a mask}: wear(John,mask). We  pivot off this \textbf{explanatory} representation to answer questions such as  \textit{What is John's stance towards mask wearing?} or \textit{What concerns does John have about mask wearing?}

Stance detection has recently (re-)emerged as a very active research area, yet many approaches generally equate stance to raw (bag-of-word) sentiment and often employ machine-learning based models requiring large amounts of (labeled or unlabeled) training data. Within such approaches, the notion of \textit{stance} varies, but generally falls into one of a handful of ``sentiment-like'' categories for \textit{stance holder} X regarding topic Y, i.e., X agrees/disagrees with Y \cite{Umer2020}, X favors/disfavors Y \cite{Krejzl2017}, X is pro/anti Y \cite{samih2021}, X has a positive/negative opinion about Y \cite{Aldayel:2021}, or X is in favor/against/neither Y \cite{10.1145/3369026}. 

We adopt the stance definition of \citet{Mather2021}, originally formulated for Covid-19.  There, a stance is a \textbf{belief-driven sentiment}, derived via propositional analysis (i.e., \textit{I believe masks do not help [and if that belief were true, I would be anti-mask]}), instead of a bag-of-words lexical matching or embedding approach that produces a basic pro/anti label.  This variant of stance detection uses a proposition to identify a domain-relevant belief in the Covid-19 domain; the belief is leveraged to compute a belief-driven sentiment and attitude toward a topic in that domain (e.g., \textit{masks}). For example, \textit{I believe masks do not protect me} is rendered as a belief type \textsc{protect} with an underlying proposition protect(masks) and a negative overall stance toward the propositional content: \textit{masks}. 

We implement and evaluate an analogous propositional framework for a new task, \textit{concern detection}. Table~\ref{tab:Examples} shows stance and concern detection output on a tweet from an English subset of a Kaggle Twitter dataset (1.9M tweets) for a new domain, the 2017 French Elections \cite{kagglefrenchelection}. The proposition common to both is ruin(jean-luc melenchon, economy). For stance, a belief \textit{type}, \textsc{destroy}, is coupled with \textit{values}: (1) Belief strength 
ranges from certainty that the belief is not true (-3) to certainty that the belief is true (+3), with 0 as ``uncommitted''; and (2) Belief-driven sentiment strength ranges from extremely negative (-1) to extremely positive (+1), with 0 as ``neutral''. 

We define a \textit{concern} to be a topical \textit{type} (e.g., \textsc{economic}) coupled with a set of \textit{values}: (1) moral dimensions from Moral Foundations Theory (MFT) \cite{haidt2004,  Graham2009, Graham2011, Graham2012}, represented as vice/virtue pairings (authority/subversion, care/harm, fairness/cheating, loyalty/betrayal, and purity/degradation); and (2) corresponding endorsement values, where ``vice'' is between 1 to 5 and ``virtue'' is $>$5 to 9 (1 is strong vice and 9 is strong virtue).
Dimensions and endorsements shown in Table~\ref{tab:Examples} are derived from a state-of-the-art baseline (Moral 1), see \S\ref{sec:task-adaptation}.

Propositional representation is the centerpiece of both stance detection and concern detection, distinguishing our lexical-based approach from model-based approaches trained on word-level annotations. Predicate-argument structure captures relationships between multi-word constituents that need not be contiguous, thus inducing \textbf{explainability}. That is, \textit{Jean-Luc Melenchon} is not adjacent to \textit{economy}, yet these terms are crucially related via the intermediate term \textit{ruin}. This enables answers to questions such as \textit{What does the author believe Jean-Luc Melenchon did to the economy?}

Moreover, domain adaptation is streamlined through predicate-argument annotation, reducing effort needed for human annotation. Annotation at the level of predicate-argument pairs factors out commonalities, reducing redundancy in the resource building process. During resource building each verb is visited only once rather than the multiple times required for word-level corpora annotation (see \S\ref{sec:task-adaptation}). For example, annotation of the word \textit{lead} is done all in one shot with a handful of automatically presented de-duplicated cases, whereas corpus-based annotation requires repeated annotations of \textit{lead}, substantially increasing human labor.

We demonstrate that the key to reduced adaptation time is the coupling of semi-automatic resource building for concern types with automated expansion of domain-independent concern values using semantic similarity. We develop a ground truth (GT) based on expert annotators and compare concern detection output to GT. We also demonstrate that, with each lexicon expansion, the performance of concern detection improves significantly over a state-of-the-art baseline moral dimension lexicon. We obtain a 231\% improvement in recall over a strong baseline for our best performing system, with only a 10\% loss in  precision.

\begin{table}
    \centering
    \begin{tabular}{|p{1.2in}|p{1.4in}|}\hline
     
      \multicolumn{2}{|c|}{\textbf{Stance Proposition:}}\\
      \multicolumn{2}{|c|}{\textit{ruin(jean-luc melenchon, economy)}}\\ \hline
      \textbf{Belief Type} & \textbf{Belief/Sent Values}\\
      \textsc{destroy} & 
      belief strength: 2.5,\newline 
      sentiment strength: -1\\ \hline \hline
      
      \multicolumn{2}{|c|}{\textbf{Concern Proposition:}}\\
      \multicolumn{2}{|c|}{\textit{ruin(jean-luc melenchon, economy)}}\\ \hline
      \textbf{Concern Type} &  \textbf{Moral Dims/Values}\\
      \textsc{economic} & care: 1.4, purity: 2.75\\ \hline
    \end{tabular}
    \caption{Representative Stance and Concern system output for a given tweet in the 2017 French Election Domain: \textit{Marine Le Pen LEADS in French poll as far left Jean-Luc Melenchon `could ruin economy'}.  Belief ranges from -3 to 3, sentiment from -1 to 1, and the moral dimensions from 1 to 9. }
    \label{tab:Examples}
\end{table}

\section{Background and Motivation}
\label{sec:background}
This section provides background and motivation for task and domain adaptability applied to 
concern detection in the 2017 French Election domain (\S\ref{sec:domain-and-task-adapt}, \S\ref{sec:stance-and-concerns}), including concern values induced from Moral Foundations Theory (MFT) (\S\ref{sec:mft-framework}).

\subsection{Task and Domain Adaptability}
\label{sec:domain-and-task-adapt}

Task adaptability and domain adaptability are two supporting areas of research for this work. Prior domain adaptation approaches, surveyed by \citet{ramponi2020}, have been applied to tasks such as sentiment analysis \cite{ben-david2020, ghosal2020}, stance detection \cite{xu2019}, and event trigger identification \cite{naik2020}. Task adaptation approaches \cite{gururangan2020, Garg2020, ziser2019} have been applied to tasks such as answer selection for question answering.

To date, both types of adaptation rely heavily on machine learning (ML) techniques, many of which require a large amount (e.g., 1M+, \citet{gururangan2020}) of training data (whether labelled or not). Some approaches employ smaller datasets, e.g., 10K+ Amazon reviews, fake news articles \cite{ben-david2020, xu2019}. Additionally, while explainability has recently been brought to the fore in deep learning approaches, as surveyed by \citet{Xie2020ExplainableDL}, such systems have not focused on task and domain adaptability.

We develop a general framework for resource building techniques that features task adaptability and retains the ability to adapt quickly to new domains. Our dataset requirements are much more minimal than prior approaches (2500 tweets per domain), there is no human labeling of corpora, and no model training is required. Moreover, explainability is achieved by virtue of inclusion of propositional information (who did what to whom) that serves as a window into the process of detecting concern types and moral dimensions.

\subsection{Stance and Concerns}
\label{sec:stance-and-concerns}

\citet{pirolli2021} apply a belief-based formulation of stance in the Covid-19 domain, with topics such as \textit{mask wearing} and \textit{social distancing}. For example, a \textit{stance} assigned to \textit{Wear a mask!} includes a \textsc{protect} belief \textit{type}, where the predicate \textit{wear} is considered a ``trigger'' and \textit{a mask} is considered the ``content'' of the belief. The \textit{values} associated with this stance include a belief strength of +3 and a sentiment strength of +1. The final \textit{stance} is thus a belief-oriented sentiment with this interpretation: the person posting the tweet is positive toward ``masks,'' assuming the belief that masks are protective is true.
In the 2017 French Election domain, a stance representation (e.g., for the example in Table~\ref{tab:Examples}) would be: 
$<$\textsc{destroy}(\textit{ruin(jean-luc melenchon, economy)}), Bel:+2.5, Sent:-1$>$.

While this prior framework lays the groundwork for domain adaptability, it has not been shown to be generalizable to new tasks (within or across domains), which is the focus of this paper. We leverage the propositional underpinnings of the framework of \citet{Mather2021} to enable a straightforward adaptation from stance detection to a new task, \textit{concern detection}, while also retaining domain adaptability. This task involves extraction of a \textit{concern type} (e.g., \textit{immigration, taxation}) associated with a given domain (e.g., \textit{2017 French Elections}), analogous to the extraction of a \textit{belief type} for a given stance detection domain.

In this paper, we demonstrate that it is straightforward to port belief-targeted stance both to a new domain (French elections) and, through an analogous proposition-based extraction, to a new task: Concern detection. 
An example of a formal Concern representation in the 2017 French Election domain for the example in Table~\ref{tab:Examples} would be: 
$<$\textsc{economic}(\textit{ruin(jean-luc melenchon, economy)}), Care: 1.4, Purity: 2.75$>$.

The approach described herein focuses on lexicon expansion obtained automatically through semantic similarity to map key terms in propositional statements to moral foundation lexicon words, using WordNet \cite{Fellbaum1998}.\footnote{Wordnet is released under a BSD style license and is freely available for research and commercial use.} Three different variants of lexicon expansion (described in \S\ref{sec:concern-value-induction}) improve on results obtained using the current state-of-the-art moral lexicon of \citet{araque2020}, which we take to be a strong baseline (henceforth referred to as `Moral 1'). The advance beyond this prior work lies in the prudent (automatic) selection of terms designated for expansion, based on propositional structure, and the combination of moral dimensions with concern types. 

\subsection{The MFT Framework and Influence}
\label{sec:mft-framework}

We focus on concern detection because identifying critical issues discussed online within a particular domain is important and useful, as is identifying the moral justifications or deliberate appeals to moral identity in these discussions.  We use the Moral Foundations Theory (MFT) framework \cite{haidt2004,Graham2009,Graham2011,Graham2012} to encode the moral dimensions of social media contributions. These moral dimensions may serve as potential indicators of influence attempts, as in \textit{When it comes to immigration it’s not about children, it’s about damaging our country!}, where the Concern type is \textsc{immigration\_refugee} and there is an appeal to the vice side (\textit{harm}) of the \textit{care/harm} moral dimension. 

An emphasis on highly controversial and/or polarizing topics in online posts/messages may be indicative of an attempt to sway others. More importantly, if these posts/messages are interwoven with language that reflects (and speaks to) the moral values of the target audience it can increase in-group cohesion, and that may further contribute to polarization. Additionally, deliberate use of morality to justify harmful intentions towards others may foster online outrage disguised as ethical conduct \cite{bandura1996mechanisms, friedman2021toward}. 

Several studies show that social groups provide a framework in which moral values are endorsed, and when these values are threatened by e.g., opposing political ideology, existing beliefs of the group are strengthened \cite{van2018partisan}. Thus, when presented with information that is incongruent with our identity and in-group, we tend to override accuracy motives in favor of social identity goals (partisan bias). When accuracy and identity goals are in conflict, moral values determine which belief to endorse and thus how to engage with information. This makes moral values an ideal breeding ground for influence campaigns, 
but also very useful for our stance and concern detection tasks. 

We note that most studies of cross-cultural values, beliefs, and morality have been conducted by WEIRD (western, educated, industrialized, rich, and democratic) countries \cite{goodwin2020cross, henrich2010most}. Inglehart's model of World Values \cite{inglehart2010wvs} has surveyed 60 countries over the last 40 years, taking into account that many nations are more concerned with economic and physical security (e.g. survival), while self-expression values are more reflective of Western countries. For example, in Pakistan or Nigeria 90\% of the population say that God is extremely important in their lives, while in Japan only 6\% take this position. 
Similarly, Schwartz' Theory of Basic Values \cite{Schwartz2012} uses a different set of organizing principles, e.g., values that relate to anxiety (e.g., tradition, security, control of threat) which may lead to an increased belief in misinformation \cite{jost2003political}.
Thus, moral dimensions combined with concern types are a potential indicator of a common actor (possibly an outside influencer) if several individuals or accounts (potentially purporting to be individuals) invoke the same moral dimensions across their messages.



\section{Task and Domain Adaptation}
\label{sec:task-adaptation}
Adaptation of stance detection to concern detection gives rise to a new framework for rapid development of a task-adapted system that retains domain adaptability and uses relatively low amounts of data, with only 2-4 hours of human categorization. 

\subsection{Resource Building Generalizations}
\label{sec:resource-building-generalizations}

We generalize to a new task while retaining domain adaptability by leveraging the stance-concern analogy, through: (a) semi-automatic domain-dependent extraction of \textit{types} from propositional analysis, i.e., moving from \textit{belief types} for stances to \textit{concern types} for concerns; and (b) fully automatic domain-independent induction of associated \textit{values} from a combination of propositional arguments and lexical and semantic resources, i.e., belief/sentiment \textit{strengths} (cf. \cite{bak:15}) for stances and moral dimensions and endorsements (cf. \cite{Graham2012}) for concerns.

Resource building for domain-dependent stance \textit{types} involves propositional analysis using semantic role labeling (SRL) \cite{gardner2018allennlp}\footnote{We use AllenNLP 2.7.0 2020 structured-prediction-srl-bert model for SRL.} to detect positions with the most highly relevant content terms, e.g., \textit{masks}. The work of \citet{Mather2021} indicates that these positions are ARG0 and ARG1. 
To port this approach over to the
induction of domain-specific concern \textit{types}, we conducted a similar analysis and found that the same positions (ARG0 and ARG1) contain the most highly relevant terms for concerns, e.g., \textit{economy}. Semi-automatic induction of concern types thus leverages these positions, as described in \S\ref{sec:concern-type-induction}.


Just as stance resource building induces domain-independent stance \textit{values} (belief / sentiment strengths), concern resource building induces domain-independent concern \textit{values} (moral dimensions / endorsements). A deeper propositional analysis reveals that additional SRL positions have a high likelihood of association with moral dimension terms, e.g., \textit{ruin}: V, ARG2, ARGM-ADV, ARGM-MNR, ARGM-PRD.
This discovery further generalizes the original stance resource-building approach and enables rapid task adaptation to concerns through entirely automatic means. We leverage these additional SRL positions to extract candidate terms for expansion of moral dimensions. Associated endorsements are then inherited from semantically similar terms from baseline Moral 1. Further details about the expansion of moral dimensions are provided in \S\ref{sec:concern-value-induction}.

Domain adaptability is retained---on analogy with stance detection---by separating and independently addressing two aspects of concern detection: (a) induction of domain-specific concern types; (b) induction of domain-independent moral dimensions. Lexicon expansion using this approach can thus be applied to domains beyond the French Elections presented herein. 

\subsection{Concern Type Induction}
\label{sec:concern-type-induction}

We adopt a generalized semi-automatic process for \textbf{lexically based} concern type induction that retains domain adaptability (later referred to as `Concern 1'). A small set (approximately 15) of domain-relevant key terms (e.g., health, taxation, immigration), is provided by a domain expert as input to a semi-automatic resource building tool.
These terms are used to filter the domain-relevant dataset. The filtered tweet subset (214k) is then divided into training (2500),\footnote{Training data are strictly for one-time semi-automatic resource building, not for model training.} and development (211,500) subsets, 
and propositional analysis is applied to the training set in a 3-step process.

First, the top 25 most frequent terms are extracted (e.g., \textit{economy)},\footnote{spaCy 3.1.0 with model en\_core\_web\_sm \cite{spacy} is used for sentence splitting and POS tagging.} ignoring functional elements such as stop words. Second, the verbs whose relevant SRL positions (defined in \S\ref{sec:resource-building-generalizations}) contain any of these top 25 terms are extracted, and the top 40 most frequent verbs e.g., \textit{ruin, restrict} are selected for further processing. Lastly, the top 10 most frequent terms associated with relevant SRL positions (for each of the 40 verbs) are extracted automatically from domain-relevant data, e.g., \textit{economy and business}, yielding 400 propositions, e.g., \textit{ruin(economy), restrict(business)}.\footnote{The thresholds of 25, 40, and 10 are selected empirically in preliminary experiments (not reported here) ascertaining a balance between adequate coverage of the data and time spent manual categorization by the expert.}

These, coupled with terms from the original domain-relevant key terms, are presented to the domain expert who constructs a small set of concern types---10 in the case of French Election: textsc{immigration\_refugee}, \textsc{electoral\_process\_voting\_laws}, \textsc{environment\_climate\_change}, \textsc{health\_care}, \textsc{taxation}, \textsc{economic}, \textsc{social\_services}, \textsc{international\_trade}, \textsc{military\_engagement}, \textsc{criminal\_justice}. Terms left uncategorized by the expert are dropped. This semi-automated concern-type induction takes 2--4 hours owing to the automatic extraction of high frequency domain relevant propositions.

\subsection{Concern Value Induction}
\label{sec:concern-value-induction}

Concern \textit{values} leverage MFT \cite{haidt2004,  Graham2009, Graham2011, Graham2012} and particularly the moralstrength library \cite{araque2020}, which serves as a strong baseline (referred to as `Moral 1').\footnote{\href{https://github.com/oaraque/moral-foundations}{https://github.com/oaraque/moral-foundations}.} This baseline lexicon includes manually developed moral dimensions (e.g. care/harm, loyalty/betrayal) and endorsement values (1--9). Our approach transcends this earlier paradigm in its application of propositional analysis with semantic-role labeling (SRL \citet{gardner2018allennlp}), coupled with a more in-depth WordNet (WN \citet{Fellbaum1998}) expansion to enrich the lexicon. This results in higher recall while retaining linguistically relevant constraints to achieve acceptable precision.

Expansion of moral dimensions relies on propositional analysis, SRL, and WordNet expansion. We select candidates for moral dimension expansion through extraction from propositional statements, and then induce three lexicon variants (in addition to the Moral 1 baseline, and a Moral 0 random chooser described in \S\ref{sec:results}) using a progression of finer-grained WordNet-based semantic-similarity functions. This expansion supports the goal of task adaptability, as required in the transition from stance detection to concern detection. 

The end result is a general approach to induction of \textit{values} for specific tasks, rendered in the form of domain-independent lexicons. That is, analogous to belief/sentiment terms of stance detection ({\em might}, {\em probably}, {\em hate}, {\em love}), we induce vice/virtue moral dimensions and their corresponding endorsement values for concern detection. 

This expansion results in three different system variations for moral dimension/values (referred to as Moral 2, Moral 3, Moral 4) beyond the Moral 1 baseline. We note that moral dimensions are assigned automatically to each lexical entry via semantic similarity to Moral 1 terms; the endorsement values are then \textit{inherited} from the most semantically similar term from the Moral 1 lexicon. An excerpt of lexicon output is shown below, from the best performing lexicon (Moral 4):\newline
\mbox{~~}{\em hypocrite} - dim: betrayal; endorse: 1.0 (strong)
\newline
\mbox{~~}{\em appreciation} - dim: care; endorse: 8.57 (strong)\newline
\noindent
\textbf{Lexicon Expansion Details:}  Moral 2-4 rely on automatic propositional analysis for prudent selection of words from the training data (via SRL, see \S\ref{sec:resource-building-generalizations}) to be considered candidates for lexicon expansion. The highest similarity match is recorded to inherit the corresponding moral endorsement value from Moral 1. A brief description of all lexicon expansions used for induction of moral dimensions and values for concern detection is provided below. (See detailed description in Appendix~\ref{sec:appendix-expansions-details} and links to Moral 2-4 lexicons in Appendix~\ref{sec:appendix-moral-lexicons}.) 

\textit{\textbf{Note:} The term ``lemma-matched'' below refers to a match between a word in a training tweet and a synset's first lemma.}
\newpage
\noindent
\textbf{Moral 1:} This initial moral dimension lexicon developed by \citet{araque2020} serves as a strong baseline, with 2800+ terms across five moral dimensions. This was derived by expanding an initial crowd sourced lexicon of about 480+ terms, annotated for moral dimensions and endorsements. Expansion to 2800+ terms was via WordNet synset matching, without regard to propositional analysis.

\noindent
\textbf{Moral 2:} (Added 214 terms, for total of 3064) This lexicon expansion yields a set of terms whose lemma-matched synsets are semantically similar (above a threshold) to lemma-matched synsets of the words in the strong baseline (Moral 1) lexicon.

\noindent
\textbf{Moral 3:} (Added 995 terms, for total of 3845) This lexicon expansion yields a set of terms whose lemma-matched synsets and their descendents are semantically similar (above a threshold) to lemma-matched synsets and their descendents of the words in the strong baseline (Moral 1) lexicon.

\noindent
\textbf{Moral 4:} (Added 5623 terms, for total of 8473) This lexicon expansion yields a set of terms drawn from \textbf{all} synsets and their descendents that are semantically similar (above a threshold) to \textbf{all} synsets and their descendents of words in the strong baseline (Moral 1) lexicon, without lemma matching.

\section{Annotation for Ground Truth}
\label{sec:annotation}

We conduct an annotation task to develop Ground Truth (GT), against which to compare our concern detection system variants, based on 50 held-out tweets from the held-out development portion of the English subset of the Kaggle Twitter dataset on 2017 French elections \cite{kagglefrenchelection}. Ground truth was produced for concern types and vice/virtue pairs for any of five moral dimensions, in accordance with guidelines in Appendix~\ref{sec:appendix-annotation-guidelines}. Annotation was completed by two non-algorithm developers (one with expertise in linguistics, the other with expertise in psychosocial moral indicators). For concern types the inter-rater reliability (IRR) is calculated through macroaveraging of kappa scores \cite{carletta-1996} which produces a 66\% agreement, considered \textbf{Strong} according to \cite{mchugh2012}. By contrast, the macroaveraged IRR for moral dimensions is low (16\%), which is considered \textbf{Weak}.

Given the high annotator reliability for concern types, system output is compared against the union of both annotators, yielding the concern type scores in Table~\ref{tab:concern_eval}. However, the lower IRR for moral dimensions is an indication that research in this realm is still in nascent stages and significant training for the task is required to achieve a reliable GT.  We note that ``ground truth'' is inherently problematic with moral indicators given the complex way in which dimensions vary with socio-cultural factors. 
Thus, for moral dimensions, system output is compared only against the single annotator with expertise in psychosocial moral indicators.

Appendix~\ref{sec:appendix-GT} presents the Ground Truth resulting from these annotations. Kaggle data \cite{kagglefrenchelection} are open and publicly available, intended for research on text analytics. The data carry no privacy or copyright restrictions. Furthermore, IRB designates our work as non human subject research. Annotators spent two hours apiece on the task. 

\section{Sample Runs}

We have implemented/validated a system to detect concerns based on induced lexical resources, using spaCy and SRL. We derive a proposition coupled with a concern \textit{type} and \textit{values} (moral dimensions/endorsements). Systems produce <concern,vice/virtue> pairs, with an average runtime of 0.6s per tweet on Mac, with no GPUs required.
\begin{table}
\begin{tabular}{|p{2.85in}|}\hline
\noindent
\textbf{Example 1}: \textit{Vatican's completely surrounded by a wall w/an entrance that's guarded 24/7 \& refugees r not allowed}\newline
\noindent
\textbf{Concern type}: \textsc{immigration\_refugee}\newline
\textbf{Proposition}: \textit{surrounded(vatican, by a wall\newline \mbox{~~}...guarded 24/7 \& refugees r not allowed))}\newline 
\textbf{Dimensions \& Endorsements}: Harm: 3.46,\newline \mbox{~~} Authority: 6.02, Degradation: 2.71\\ \hline
\noindent
\textbf{Example 2}: \textit{Where is the justice?}\newline
\noindent
\textbf{Concern type}: \textsc{criminal\_justice}\newline
\textbf{Proposition}: \textit{is(where, the justice)}\newline
\textbf{Dimensions \& Endorsements}: Fairness: 7.6\\ \hline
\noindent
\textbf{Example 3}: \textit{Most Melanchon voters care more about their country than their ideological purity.}\newline
\noindent
\textbf{Concern type}: \textsc{electoral\_process\_ \newline \mbox{~~}voting\_laws}\newline
\textbf{Proposition}: \textit{care(most melanchon voters, \newline \mbox{~~} more about their country)}\newline
\textbf{Dimensions \& Endorsements}: Care: 8.80, \newline \mbox{~~}Loyalty: 7.14, Authority: 5.57,\newline \mbox{~~}Degradation: 3.69\\ \hline
\end{tabular}
\caption{Sample Output from Concern Detection on Kaggle 2017 French Elections; SRL output is reproduced including any errors}
\label{tab:sample-concern}
\end{table}

Representative outputs on a 1K-tweet held-out portion of the development dataset from Kaggle 2017 French Elections are shown in Table~\ref{tab:sample-concern}. These are taken from our best performing variant described in \S\ref{sec:results}. In Example 1, \textsc{immigration\_refugee} (triggered by \textit{refugees}) is coupled with moral dimension/endorsement \textit{values}, Harm (\textit{wall, entrance, guard}), Authority (\textit{wall, guard}), and Degradation (\textit{entrance}). In Example 2, \textsc{criminal\_justice} (triggered by \textit{justice}) is coupled with moral dimension/endorsement \textit{values}: Fairness (\textit{justice}).  Both are reasonable outputs.

In example 3, \textsc{electoral\_process\_} \linebreak \textsc{voting\_laws} (triggered by \textit{voters}) is coupled with moral dimension/endorsement \textit{values}, Care (\textit{care}), Loyalty (\textit{country}), Authority (\textit{care, country}), and Degradation (\textit{care, voters}).  This example overgenerates, assigning Degradation based on the terms \textit{care and voters}. A further lexicon enhancement is needed (as alluded to in \S\ref{sec:conclusion-future-work}) for elimination of spurious lexical entries that lead to false positives (i.e., a reduction in Precision).

\section{Results and Analysis}
\label{sec:results}

We explore the performance of each of our four moral dimension lexicons by comparing $<$concern,vice/virtue$>$ outputs against GT for each lexicon. Concerns types are evaluated for an exact match against the GT concern (e.g., \textsc{criminal\_justice}) and moral values are evaluated for an exact match against the binary choice in the GT (vice or virtue). Concern 1 represents the lexicon-based chooser described in \S\ref{sec:concern-type-induction}. Our (strong) baseline for moral values is ``Moral 1'' \cite{araque2020}. Subsequent variants (Moral 2-4) are the expanded moral dimension lexicons using the techniques described in \S\ref{sec:concern-value-induction}. 
We also compare against random choosers for Concern \textit{Types} (Concern 0) and \textit{Values} (Moral 0).  Table~\ref{tab:concern_eval} shows the results of each system output compared to GT.
\begin{table}[ht]
    \centering
    \footnotesize
    \begin{tabular}
    {|p{0.52in} 
     ||p{.19in} 
     |p{.19in} 
     |p{.19in}
     ||p{.23in}
     |p{.23in} 
     |p{.23in} 
     |} \cline{1-7}
     \bf System &
     \bf TP &
     \bf FP &
     \bf FN &
     \bf P &
     \bf R &
     \bf F1 
     \\ \hline \hline
     Concern 0&
     35 & 233 & 25 & 12.69 & 53.88 & 20.54 \\ \hline 
     Concern 1&
     \textbf{40} &
     \textbf{1} &
     \textbf{20} &
     \textbf{91.60} &
     \textbf{66.67} &
     \textbf{77.17}     
     \\ \hline \hline
     Moral 0&
     33 & 225 & 33 & 12.74 & 43.01 & 19.65 \\ \hline
     Moral 1 &
     13 &
     \bf 28 &
     53 &
     \bf 32.99 &
     19.69 &
     24.67
     \\ \hline
    Moral 2 &
     17 &
     58 &
     49 &
     31.20 &
     25.76 &
     28.22
     \\ \hline
    Moral 3 &
     24 &
     122 &
     42 &
     31.38 &
     36.36 &
     33.69
     \\ \hline
    Moral 4 &
     \bf 43 &
     181 &
     \bf 23 &
     29.75 &
     \bf 65.15 &
     \bf 40.85
    \\ \hline
    \end{tabular}
    \caption{Evaluation of (Domain-Dependent) Concern Types and (Domain-Indepenent) Moral Dimensions: Unweighted sum of true positives (TP), false positives (FP), and false negatives (FN) across all labels. Precision (P), recall (R), and F1 numbers for all systems are weighted macro-averages except for the random choosers (Concern 0 and Moral 0). Lexicon variants affect only the moral dimensions.}
    \label{tab:concern_eval}
\end{table}

System performance is measured by weighted macro-averaged precision (P), recall (R), and F1 scores.
Domain-dependent Concern \textit{Type} is not affected by moral lexicon variants and independently has its own P/R/F1 scores.  Concern \textit{Values} (i.e., moral Dimensions) are impacted by lexicon variants and therefore have a row corresponding to each variant.
We note the importance of applying a \textbf{weighted} macro average to these scores due an imbalance in the distribution of classes \cite{delgado2019}, where the probability of one class can be substantially higher or lower from others. For example, we observed that the number of \textsc{electoral\_process\_voting\_laws} annotations is 2.3 times higher than the number of \textsc{international\_trade} annotations.

One might expect random choosers (Concern 0 and Moral 0) to have a decent likelihood of getting many hits, with an expected rate of about 50\% that each Concern type and Moral value will be selected. If so, this would result in an expected 250 positive selections for concern types (whereas the two annotators together only made 60 positive selections) and an expected 250 positive selections for moral values (whereas the expert annotator only made 66 positive selections). However, the results in Table~\ref{tab:concern_eval} indicate that, while the increased number of hits leads to a reasonably high recall, the number of false positives (233) swamps out the hit rate, leading to a low F1 score (20.54). Accordingly, Concern 1 easily beats the random choice baseline by a healthy margin, with an F1 score of 77.17.


In contrast, Moral 0 achieves 79.65\% of the performance of the Moral 1 baseline, with an F1 score of 19.65---not too far off from the 24.67 baseline F1 score. Moreover, the F1 scores for Moral 2-4 surpass this baseline, with statistically significant improvements indicated between all system pairs at the 3.5\% level or better, according to the McNemar statistical test \cite{McNemar47samplingIndependence}.\footnote{Tested values are correct responses (TP or TN) vs. incorrect responses (FP or FN), for determining significance of change in total error rate.} That is, all lexicon expansion improvements are statistically significant. Notably, a 231\% improvement in Recall is achieved for Moral 4 over the strong baseline (Moral 1): 65.15 vs. 19.69. This is achieved with only a 10\% loss of precision, ultimately yielding an F1 score of 40.85 which is a 66\% improvement. 
\newpage
We observe that the precision-recall gap increases as the notion of similarity is loosened: (1) \textit{lead} is similar to \textit{strip} in Moral 4, but not in Moral 3 (reducing Moral 4 precision); (2) \textit{price} is similar to \textit{value} in Moral 4 but not in Moral 3 (increasing Moral 4 recall). We further assess system performance by comparing Concern 1 performance to human performance.
Average F1 score for the two annotators is 78.88, and Concern 1 performance is 77.17 F1. Concern detection thus yields 97.8\% of human performance on concern type detection.

Error analysis of FP/FN's for concern \textit{types} reveals that concern detection fails to assign any concern type (FN) to \textit{Infighting among left-wing could hand Front National VICTORY}, which is annotated as \textsc{electoral\_process\_voting\_laws}, because terms like \textit{Front} and \textit{left-wing} are not present in the concern type lexicon. 

For concern \textit{values}, the annotator does not assign a moral dimension to \textit{Macron is center right}, yet Moral 4 inaccurately detects (FP) Care, Authority, and Betrayal due to the existence of the word \textit{center} and also detects Purity from the word \textit{right}. For this same sentence Moral 1 also inaccurately detects Fairness from the word \textit{right}. Many cases similar to these impact precision values for each Moral 1-4, potentially requiring lexicon tuning (see \S\ref{sec:conclusion-future-work}).

We conduct further analysis to determine whether performance is impacted by potential overfitting of concern value detection to the domain of interest during development. We compare Araque's original concern value detection tools \cite{moralrepo}, a unigram model trained on Hurricane Sandy data from the Moral Foundations Twitter Corpus (MFTC) \cite{hoover2019}, to our proposition-based concern value detection that uses semantically expanded lexicons based on the Kaggle French elections data. We level the playing field by applying both approaches to both datasets, measuring each against their respective GTs. 

\begin{table}[ht]
    \centering
    \footnotesize
    \begin{tabular}
    {|p{0.38in} 
     |p{.78in}
     |p{.26in}
     |p{.26in} 
     |p{.27in} 
     |} \hline
     \multicolumn{2}{|c|}{\bf Kaggle 2017 French Elections} &
     \bf P &
     \bf R &
     \bf F1 
     \\ \hline
     \multirow{3}{*}{Moral 1} & 
     Proposition~(\S\ref{sec:concern-value-induction}) & 32.99 & 19.69 & 24.67 \\ \cline{2-5}
     & Unigram Model \cite{moralrepo} & 29.07 & 31.82 & 30.38 \\ \hline
     \multirow{3}{*}{Moral 4} & 
     Proposition~(\S\ref{sec:concern-value-induction}) & 29.75 & 65.15 & 40.85 \\ \cline{2-5}
     & Unigram Model \cite{moralrepo} & 20.88 & 24.24 & 22.43 \\ \hline
\multicolumn{5}{c}{\mbox{~~~}} \\ \hline   
     \multicolumn{2}{|c|}{\bf MFTC Hurricane Sandy} &
     \bf P &
     \bf R &
     \bf F1 
     \\ \hline
     \multirow{3}{*}{Moral 1} & 
     Proposition~(\S\ref{sec:concern-type-induction}) &  65.58 & 17.56 & 27.68 \\ \cline{2-5}
     & Unigram Model \cite{moralrepo} & 56.02 & 65.89 & 60.55 \\ \hline
     \multirow{3}{*}{Moral 4} & 
     Proposition~(\S\ref{sec:concern-value-induction}) & 26.10 & 35.29 & 29.99 \\ \cline{2-5}
     & Unigram Model \cite{moralrepo} & 56.02 & 65.89 & 60.55 \\ \hline
    \end{tabular}
\caption{Comparison of two Concern Value (Moral Dimensions) detection algorithms across two domains (Kaggle 2017 French Elections and MFTC Hurricane Sandy) using Moral 1 and Moral 4 lexicons. 
}
\label{tab:combined_domain_moral_dim_comparison}
\end{table}

\citet{moralrepo}'s Unigram Model performs best on Hurricane Sandy test data (Table~\ref{tab:combined_domain_moral_dim_comparison}), with weighted macroaverage F1, using both Moral 1 and Moral 4 lexicons (60.55), but does not perform as well on Kaggle test data (30.38 and 22.43). This is a potential indicator of overfitting to the Hurricane Sandy data during training. Similarly, proposition-based concern value detection (\S\ref{sec:concern-value-induction}) performs better on Kaggle test data (40.85) than on Hurricane Sandy Data (29.99)---a sign of overfitting in the opposite direction during lexicon expansion.

\begin{table}[ht]
    \centering
    \footnotesize
    \begin{tabular}
    {|p{0.4in} 
     |p{.72in}
     |p{.26in}
     |p{.26in} 
     |p{.27in} 
     |} \hline
     \multicolumn{2}{|c|}{\bf Kaggle 2017 French Elections} &
     \bf P &
     \bf R &
     \bf F1 
     \\ \hline
     Moral 1 & Full Input Text & \textbf{34.05} & 21.21 & 26.14 \\ \hline
     Moral 4 & Full Input Text & 32.42 & \textbf{62.12} & \textbf{42.61} \\  \hline
     \multicolumn{5}{c}{\mbox{~~~}} \\ \hline   
     \multicolumn{2}{|c|}{\bf MFTC Hurricane Sandy} &
     \bf P &
     \bf R &
     \bf F1 
     \\ \hline
     Moral 1 & Full Input Text & \textbf{67.23} & 22.91 & 34.13 \\ \hline
     Moral 4 & Full Input Text & 36.79 & \textbf{49.08} & \textbf{42.03} \\  \hline
    \end{tabular}
\caption{Test for Domain Stability of Concern Value (Moral Dimensions) detection across two domains (Kaggle 2017 French Elections and MFTC Hurricane Sandy), using Moral 1 and Moral 4 lexicons on full text input. 
}
\label{tab:full-input-text}
\end{table}

Testing our moral lexicon expansion for stability retention across domains, we level the playing field by relaxing the constraint that only a propositional sub-portion of the text is considered as input. We apply our algorithm to the entire text (full tweet) to induce a fairer comparison with Araque et al's Unigram model. Table~\ref{tab:full-input-text} shows comparable or better macroaveraged F1 scores than those shown for ``Proposition'' in Table~\ref{tab:combined_domain_moral_dim_comparison}.

More importantly, our expanded Moral 4 lexicon demonstrates stability across domains: 42.61 for 2017 French Elections and 42.03 for Hurricane Sandy (vs. 26.14 and 34.13, respectively, for Araque et al's Moral 1 lexicon). This illustrates the potential for proposition-based \textbf{domain adaptability} but highlights the need for hybridized detection to ensure both \textbf{task adaptability} and \textbf{explainability} (discussed further below). 

\section{Conclusions, Limitations, Future Work}
\label{sec:conclusion-future-work}

We implement and validate a generalized paradigm for adaptation of tasks involving propositions to a new task (concern detection) and domain (French elections). Our primary contribution is the provision of a framework for rapid adaptability, based on: (a) semi-automatic domain-dependent extraction of \textit{types} from propositional analysis; and (b) fully automatic domain-independent induction of \textit{values} from propositional arguments and semantic resources. We demonstrate that the coupling of (a) and (b) leads to rapid ramp-up resource construction for a new task and domain (2-4 hours). 

We demonstrate that a deeper propositional analysis is key to generalizing domain-adaptable resource-building for new tasks. We develop an automatic procedure for expanding moral dimensions that incorporates propositional analysis, semantic-role labeling, and in-depth WordNet (WN) expansion, to produce three increasingly expanded moral dimension/endorsement lexicons. We develop a ground truth (GT) based on expert annotators and compare our concern detection output to GT, to yield 231\% improvement in recall over baseline, with only a 10\% loss in precision. F1 yields 66\% improvement over baseline and 97.8\% of human performance. Our 
approach yields large savings over those that employ costly human labor and model building. Work produced herein provides the community with a 
newly expanded moral dimension/value lexicon, annotation guidelines, and GT for 50 tweets, intended for research purposes.

We show that our proposition-based moral lexicon expansion provides stability across domains. However, the results of concern value detection using the full text of an input (a tweet) highlight the importance of adopting a hybrid approach that captures fine-grained distinctions and produces an explainable representation that is not otherwise available (e.g., in the unigram language model of \citet{moralrepo}). An avenue of future research is to explore a hybridized approach that makes fine-grained concern value distinctions for \textit{domain adaptability}, while leveraging an explainable propositional representation for \textit{task adaptability}. As a first step, we will apply our propositional analysis iteratively on full-text inputs to detect answers to questions not otherwise extractable from raw textual strings (see related discussion in \S\ref{sec:intro}).

Our results do not require any tuning of the lexicons to remove terms that result in a high number of false positives (FP) and false negatives (FN). Future work will explore fine tuning of the lexicons to address cases seen in \S\ref{sec:results} with an eye toward improving the precision without a large drop in recall, to yield an even higher F1-score. 


A current limitation (for future study) is the omission of additional moral dimensions, e.g., liberty/oppression \cite{haidt2012righteous}.
This reflects political equality related to dislike of oppression and concern for victims, not a desire for reciprocity. 
In political discourse, this is apparent in anti-authoritarianism and
anti-government anger,
which makes
it an important dimension for the topic of the French election \cite{ravi2012}
and widens the range of
potentially relevant information that could indicate influence attempts. 
 
Future work also includes exploration of other cultural frameworks, in addition to or in place of Moral Foundations Theory, e.g., Inglehart’s Cultural Map model \cite{inglehart2010wvs} and Schwartz Value Theory \cite{Schwartz2012}. Cultural models that allow more room for non-Western values (e.g. survival needs) are important for reducing bias, and a feasible avenue for improving the performance and applicability of concern detection.

Another limitation is that GT development for a given cultural context is difficult, especially with diverse annotators.
\citet{prabhakaran2021releasing} show that systematic disagreements between annotators with differing socio-cultural backgrounds are obfuscated through aggregating crowd-sourced annotations. Future work will explore vector-based approaches to assign weights based on their representativeness in a given culture. 

Cultural values are also reflected in language (e.g., gendered vs. non-gendered languages, culturally intrinsic concepts).
Accordingly, 
our future work involves processing concerns for other languages through adaption of SRL \cite{gardner2018allennlp} to multilingual input, starting with French, employing multilingual preprocessing via spaCy \cite{spacy} and EuroWordNet and related multilingual WordNets \cite{bond2012, bond-foster-2013-linking}, to expand moral dimensions to other languages.


\section*{Ethical Considerations} Annotation was completed by two non-algorithm developers (one with expertise in linguistics, the other with expertise in psychosocial moral indicators), compensated appropriately for their work.  A two-level institutional review board (at both the sponsored site and at the sponsoring site) deemed this work as ``research not involving human subjects,'' as it does not involve a living individual about whom an investigator conducting research obtains information through intervention or interaction with the individual, or obtains, uses, studies, analyzes, or generates identifiable private information. 

The data used for the software development is provided from Kaggle's existing public research Twitter dataset, focusing on an English subset for the 2017 French Elections of 1.9 million tweets \cite{kagglefrenchelection} at \url{https://www.kaggle.com/jeanmidef/french-presidential-election}. Kaggle data contain user names, but the dataset is open and publicly available.

Potential risks may emerge from language biases in standard resources on which some of our work is built. For example, cultural idioms like \textit{fluctuat nec mergitur} may be translated from Latin into the correct literal meaning of \textit{tossed [by the waves], but not sunk}, but the culturally distinct values (Paris' coat of arms and motto with a deep affective history) will get lost in translations into English, and with it, its cultural meaning. Similarly, while English WordNet provides one of the most comprehensive
semantic ontology of words in English,
embedded biases are still present today, e.g., 
offensive, racist, and misogynistic slurs \cite{crawford2021excavating}. These issues need to be addressed within the language resource community.

Risk of misuse of this technology is mitigated by the transparent nature of concern detection, owing to the propositional representations that underlie and inform algorithmic decisions. In contrast to ML approaches, 
misuse within the technology would be easily discoverable. The technology further serves as a framework within which cultural distinctions may be studied and better understood, thus mitigating the potential for cross-culturally undetected misuse.

\section*{Acknowledgments}

This material is based upon work supported by the Defense Advanced Research Projects Agency (DARPA) under Contracts No.~HR001121C0186, No~HR001120C0037, and PR No.~HR0011154158. Any opinions, findings and conclusions or recommendations expressed in this material are those of the authors and do not necessarily reflect the views of the Defense Advanced Research Projects Agency (DARPA).

\newpage

\bibliographystyle{acl_natbib}

\appendix
\section{Lexicon Expansion Detailed Descriptions}
\label{sec:appendix-expansions-details}

Below is the detailed description of each expansion algorithm.
\noindent
\textbf{Basic definitions:}\\
\begin{math}
w = \mbox{word in training tweet (e.g., `concerned')}\\
m = \mbox{moral foundations word (e.g., `concern')}\\
wn = \mbox{wordnet package}\\
S_{w} = \mbox{wn.synsets(w) (set of synsets for word w)}\\
s_{w,1} = \mbox{wn.synsets(w)[0] (1st synset for word w)}\\
L_{i} =  \mbox{lemmas associated with } S_{i}\\
l_{i,k} = \mbox{s.lemmas() (lemmas for synset } s_{i})\\
l_{i,1}=\mbox{s.lemmas()[0].name() (1st lemma synset } s_{i})\\
\end{math}
\textbf{Note:} Use of the term ``lemma-matched'' below refers to a match between a word in a training tweet and a synset's first lemma.

\noindent
\textbf{Lexicon Expansion:} All lexicon expansions (Moral 2-4 below) beyond a strong baseline (Moral 1) rely on propositional guidance to select words from the training data to be considered candidates for lexicon expansion. The highest similarity match is recorded for future selection of the moral endorsement value. All moral lexicon expansions (Moral 2-4 below) beyond a strong baseline (Moral 1) apply to each pair $(w, m)$ for each word of interest $w$ and each moral foundations word $m$ ($w\times m$ iterations):

\noindent
\textbf{Moral 1:} This initial moral dimension lexicon developed by \citet{araque2020}\footnote{\href{https://github.com/oaraque/moral-foundations}{https://github.com/oaraque/moral-foundations}.} serves as a strong baseline, with 2800+ terms across five moral dimensions. This was derived by expanding an initial crowd sourced lexicon of about 480+ terms, annotated for moral dimensions and endorsements. Expansion to 2800+ terms was via WordNet synset matching, without regard to propositional analysis.

\noindent
\textbf{Moral 2:} (Added 214 terms) This lexicon expansion yields a set of terms whose lemma-matched synsets are semantically similar (above a threshold) to lemma-matched synsets of the words in the strong baseline (Moral 1) lexicon, using the following steps: (a) Extract all synsets $s_{w,i}$  in $S_w$ (for word $w$) whose first lemma $l_{i,1}$ exactly matches $w$, producing $S_x$.(b) Extract all $s_{m,j}$ in $S_m$ (for word $m$) whose first lemma $l_{j,1}$ exactly matches $m$, producing $S_y$. (c) Return all lemmas $l_{k,1}$ of any $s_{x,k}$ in $S_x$ that matches any $s_{y,h}$ in $S_y$ using wordnet wup\_similarity  w/ threshold 0.9 (if any).
Example: concerned → concerned.a.01.

\noindent
\textbf{Moral 3:} (Added 995 terms) This lexicon expansion yields a set of terms whose lemma-matched synsets and their descendents are semantically similar (above a threshold) to lemma-matched synsets and their descendents of the words in the strong baseline (Moral 1) lexicon, using the following steps: (a) Extract all synsets $s_{w,i}$, in $S_w$ (for word $w$) whose first lemma $l_{i,1}$ matches exactly $w$, producing $S_x$. Extract all synsets $s_{m,j}$ in $S_m$ (for word $m$) whose first lemma $l_{j,1}$ matches exactly $m$, producing $S_y$. (First part of 2 up to this point.)  (b) Expand lemmas from both sets: (i) collect all lemmas for all synsets in $S_x$, producing $L_x$; (ii) collect all lemmas for all synsets in $S_y$, producing $L_y$. (c) Extract synsets for these lemma expansions: (i) collect all synsets for all lemmas in $L_x$, producing $S_a$; (ii) collect all synsets for all lemmas in $L_y$, producing $S_b$. (d) Return all unique lemmas $l_{c,1}$ of any synset $s_{a,c}$ in $S_a$ that matches a synset $s_{b,d}$ in $S_b$ using wup\_similarity w/ threshold 0.9 (if any).
Example: concerned → concerned.a.01 → implicated → implicated.s.01

\noindent
\textbf{Moral 4:} (Added 5623 terms) This lexicon expansion yields a set of terms drawn from \textbf{all} synsets and their descendents that are semantically similar (above a threshold) to synsets and their descendents of words in the strong baseline (Moral 1) lexicon---without lemma matching---using the following steps: (a) Extract all synsets $s_{w,i}$ in $S_w$ (for word $w$), producing $S_x$. Extract all $s_{m,j}$ in $S_m$ (for word $m$), producing $S_y$. (b) Expand lemmas from both sets: (i) collect all lemmas for all synsets in $S_x$, producing $L_x$; (ii) collect all lemmas for all synsets in $S_y$, producing $L_y$. (c) Extract synsets for these lemma expansions: (i) collect all synsets for all lemmas in $L_x$, producing $S_a$; (ii) collect all synsets for all lemmas in $L_y$, producing $S_b$. (d) Return all unique lemmas $l_{c,1}$ of any synset $s_{a,c}$ in $S_a$ that matches a synset $s_{b,d}$ in $S_b$ using wup\_similarity w/ threshold 0.9 (if any)
Example: concerned → refer.v.02 → refer → pertain.v.02

\section{Moral Dimension/Value Lexicons: Moral 2-4}
\label{sec:appendix-moral-lexicons}

Three moral lexicons are induced automatically via propositional analysis, SRL, and semantic lexicon expansion:
\begin{itemize}
    \item Moral 2: \href{https://github.com/ihmc/Findings-of-ACL-2022-Concern-Detection/tree/main/Moral\%202}{Moral 2 Lexicon}
    \item Moral 3: \href{https://github.com/ihmc/Findings-of-ACL-2022-Concern-Detection/tree/main/Moral\%203}{Moral 3 Lexicon}
    \item Moral 4: \href{https://github.com/ihmc/Findings-of-ACL-2022-Concern-Detection/tree/main/Moral\%204}{Moral 4 Lexicon}
\end{itemize}

\section{Annotation Guidelines}
\label{sec:appendix-annotation-guidelines}
Steps below refer to column labels in the Blank Annotation Sheet found  \href{https://github.com/ihmc/Findings-of-ACL-2022-Concern-Detection/blob/main/Blank\%20Annotation\%20Sheet.xlsx}{\textbf{here}}.
\begin{enumerate}
    \item For each tweet excerpt in the “Text” column A, apply steps 2--8 below.
    \item Columns B through K are the potential Concern types. Put a ``1'' in a single column corresponding to the applicable Concern type. Leave empty if none appears to apply.
    \item Columns L through W are moral dimensions. Put a ``1'' in any column that has an applicable moral dimension, in either the vice subcolumn or the virtue subcolumn. Leave the vice/virtue cells empty for any moral dimension that is not applicable. Refer to the moral dimensions described in the Graham and Haidt tables for this task (see links in 4a and 4b below).
    \item Provide annotations only for explicitly represented material. Do not infer context that is not stated and do not apply any subject-matter background knowledge. The only background knowledge to be used for this task is the moral dimensions description in tables below from:
    \begin{enumerate}
        \item \href{https://citeseerx.ist.psu.edu/viewdoc/download?doi=10.1.1.459.9448&rep=rep1&type=pdf}{[Graham and Haidt, 2012]} (see page 16)
        \item \href{https://cpb-us-e2.wpmucdn.com/sites.uci.edu/dist/1/863/files/2020/06/Graham-et-al-2013.AESP\_.pdf}{[Graham et al., 2013]} (see page 68)
    \end{enumerate}
    \item Do not try to force the Text into a particular moral dimension. If a moral dimension appears to be applicable, but it is unclear whether the vice or virtue is active, put “DK” in either the vice cell or virtue cell, instead of leaving it blank.
    \item Consider only the content of the Text without regard to grammaticality or punctuation.
    \item Assume no sarcasm is present. Annotate the literal sense. 
    \item Use the last column (optionally) for any annotation notes, eg, the reasoning behind the chosen annotations.
\end{enumerate}

\section{Ground Truth}
\label{sec:appendix-GT}

We conduct an annotation task to develop Ground Truth (GT), against which to compare our concern detection system variants, based on 50 held-out tweets from the held-out development portion of the Kaggle English Twitter dataset on 2017 French elections \cite{kagglefrenchelection}.
Annotation was completed by two non-algorithm developers (one with expertise in linguistics, the other with expertise in psychosocial moral indicators). Annotators were provided guidelines (Appendix~\ref{sec:appendix-annotation-guidelines}) for both concern types and moral dimensions. For concern types the interrater reliability (IRR) is calculated through macroaveraging of kappa scores \cite{carletta-1996} which produced a 66\% agreement, considered \textbf{Strong} according to \cite{mchugh2012}. By contrast, the macroaveraged IRR for moral dimensions is considered none to slight (16\%), which is deemed \textbf{Weak}.  [\href{https://github.com/ihmc/Findings-of-ACL-2022-Concern-Detection/blob/main/Concern\%20Detection\%20Ground\%20Truth.xlsx}{Link to GT}]

\end{document}